\documentclass[11pt,a4paper]{article}
\usepackage[T1]{fontenc}
\usepackage[utf8]{inputenc}
\usepackage{lmodern}
\usepackage[margin=2.5cm]{geometry}
\usepackage{microtype}
\usepackage{amsmath,amssymb}
\usepackage{graphicx}
\usepackage{booktabs}
\usepackage{tabularx}
\usepackage{enumitem}
\usepackage[hidelinks]{hyperref}
\usepackage{xurl}
\setlist{itemsep=2pt,topsep=4pt}

\newcommand{\pifive}{$\pi_{0.5}$}
\newcommand{\x}{$\times$}

\title{Camera equalisation of an 11.5$\times$ drop in environmental illuminance:\\ four imitation-learning policies on an SO-101 arm}
\author{Simon Cowling\\ fysical.ai, Odense, Denmark\\ \href{mailto:simon@fysical.ai}{\texttt{simon@fysical.ai}}}
\date{November 2026}

\begin{document}
\maketitle

\begin{abstract}
\noindent Real-robot benchmarks of learned visuomotor policies often record room illuminance as the lighting condition. We show that, on a typical low-cost rig, room light and image light can be almost unrelated. Four imitation-learning policies (ACT, SmolVLA, Diffusion Policy and \pifive) were trained on the same 50 demonstrations of a pick-and-place task on an SO-101 arm. Each was run for ten rollouts at about 520\,lux and again at about 45\,lux, an 11.5\x{} reduction. Success was 20/40 in normal light and 19/40 in low light: a change of $-2.5$ percentage points (95\% CI $-23$ to $+19$), with no evidence that the policies differed in sensitivity (interaction $p = 0.86$). The recorded video shows why: mean image brightness was only 3\% lower in low light on the front camera and 19\% lower on the wrist camera. Both cameras were set to automatic exposure and white balance, which most likely absorbed the difference; exposure was not locked in this experiment. An accidental repeat of one 10-rollout block reproduced its success count (7/10) but matched the first block at only 6 of 10 placements, consistent with chance. We conclude that lux alone does not describe what a policy sees. Camera settings, and image statistics compared with the training recordings, are the quantities to report and control. Data and code are public.
\end{abstract}

\section{Introduction}
This note asks how much the success of four learned manipulation policies changes when the room light drops about tenfold, and why.

Visuomotor policies map camera images to robot actions. Lighting changes those images everywhere at once, and a simulated perturbation benchmark ranks lighting among the changes that hurt manipulation policies most~\cite{colosseum}. A natural safeguard in real-robot evaluation is to fix the light and log illuminance in lux, assuming that equal lux means equal inputs.

We tested that assumption on a common low-cost setup: an SO-101 arm~\cite{so101} driven through LeRobot~\cite{lerobot}, with two off-the-shelf USB cameras. Four widely used policies were trained on identical demonstrations: ACT~\cite{act}, SmolVLA~\cite{smolvla}, Diffusion Policy (DP)~\cite{dp} and \pifive~\cite{pi05}. We then benched them at the recording light level and at about one eleventh of it. The contributions are:
\begin{enumerate}
\item \textbf{A paired real-robot comparison} of four policies under an 11.5\x{} illuminance drop: 80 rollouts, with effect estimates and intervals.
\item \textbf{Measurements of what the cameras actually recorded.} The images were barely darker, so the finding concerns the camera pipeline more than the policies.
\item \textbf{A same-day repeat of one block}, which shows how much a single 10-rollout block varies on its own.
\item \textbf{Practical recommendations} for reporting and controlling lighting in small real-robot benchmarks.
\end{enumerate}
The experiment ran as an exploratory side study during a larger, pre-registered benchmark on the same rig. It was not itself pre-registered.

\section{Setup}
The design compares the same four checkpoints, the same ten ball placements and the same settings under two light levels.

\paragraph{Rig and task.} A 6-DoF SO-101 follower arm runs LeRobot, pinned at commit \texttt{ba83e4cf}, on one workstation with an RTX 5070 Ti. Two UVC cameras are read through OpenCV at 640\x480 and 30\,fps: a fixed front camera (Logitech Brio 300) and a wrist camera (a generic USB 2.0 module). The task is: \emph{pick up the green ball and place it in the basket, ignoring the two red distractor balls.} A rollout succeeds if the green ball is in the basket within 30\,s.

\paragraph{Models.} All four policies were trained on the same 50 teleoperated demonstrations, recorded at about 500\,lux, using LeRobot's reference configurations with the schedules in Table~\ref{tab:models}. ACT's ResNet-18 starts from ImageNet weights (the LeRobot default), DP's ResNet-18 is trained from scratch, and SmolVLA and \pifive{} are fine-tuned from their pretrained vision-language-action base models.

\begin{table}[h]
\centering\small
\caption{Checkpoints benched. Cooldowns used a $1-\sqrt{\cdot}$ learning-rate decay.}
\label{tab:models}
\begin{tabular}{llll}
\toprule
Policy & Checkpoint benched & Weights & Inference \\
\midrule
ACT~\cite{act} & 30{,}000 steps, constant learning rate & raw & fp32 \\
SmolVLA~\cite{smolvla} & 6{,}000 steps (cooldown from 4{,}800) & raw & bf16 compute \\
Diffusion Policy~\cite{dp} & 50{,}000 steps (cooldown from 40{,}000) & EMA & fp32 \\
\pifive~\cite{pi05} & 6{,}000 steps (cooldown from 4{,}800) & EMA & bf16 \\
\bottomrule
\end{tabular}
\end{table}

\paragraph{Lighting conditions.} \emph{Normal} (5 Nov 2026): lux-matched to the recordings, with the room's artificial lights on and daylight unscreened. \emph{Low} (6 Nov 2026): every artificial light off and the daylight reduced with window screens and blockers. The low condition therefore changed the light's source, colour and direction as well as its level. Illuminance was read once per block at six points with a handheld lux meter; the workspace reading is the primary measure.

\paragraph{Procedure.} Each block was ten rollouts, one per fixed placement N00--N09, run in the same policy order on both days with \texttt{lerobot-rollout}. Blocks started at 09:42, 09:54, 10:06 and 11:18 on 5 Nov (ACT, SmolVLA, DP, \pifive), and at 09:30, 09:42, 09:53 and 10:52 on 6 Nov. Settings were episodic, synchronous inference, 30\,s per rollout, 20\,s reset. The command lines differed between conditions only in the output dataset name. One unblinded operator placed the ball and scored each rollout live: success (0/1); started (the arm moved purposefully toward the ball); and, for failures, one label: location, grip, ball knocked out of reach, or wrong object.

\paragraph{Analysis.} Per policy: risk differences with Newcombe 95\% intervals~\cite{wilson,newcombe}, Fisher exact tests, and exact McNemar tests over placements. Pooled: a Mantel--Haenszel odds ratio stratified by policy, with a Breslow--Day homogeneity test~\cite{mh,bd}, and binomial GLMs with policy fixed effects, extended with a policy \x{} condition interaction and with placement fixed effects. Bayesian: Beta(1,1) posteriors per cell. Images: every frame of every rollout was decoded and measured for brightness (Rec.~601 luma), contrast, sharpness (Laplacian variance on 4\x-downsampled luma, also divided by mean luma squared to remove the effect of brightness), saturation and white balance, and compared with 20 training episodes. Timestamps in the recorded datasets are nominal, so frame-rate drops were detected as repeated frames while the scene was moving. The analysis is exploratory, and $p$-values are descriptive.

\section{Results}
\subsection{Light achieved}
Workspace illuminance averaged 519\,lux (block range 477--556) in normal light and 45\,lux (41--51) in low light, a ratio of 11.5\x. Across sensor positions and block pairs the ratio ran from 7.7\x{} to 13.6\x{} (Appendix, Table~\ref{tab:lux}). Workspace light generally rose through each morning, consistent with rising daylight: 477, 538, 506 and 556\,lux in normal light; 41, 45 and 51\,lux for ACT, SmolVLA and DP in low light. The DP repeat block read 59\,lux and the low-light \pifive{} block 43\,lux. On both days \pifive{} ran last, about an hour after DP.

\subsection{Success}
\begin{figure}[h]
\centering
\includegraphics[width=0.92\linewidth]{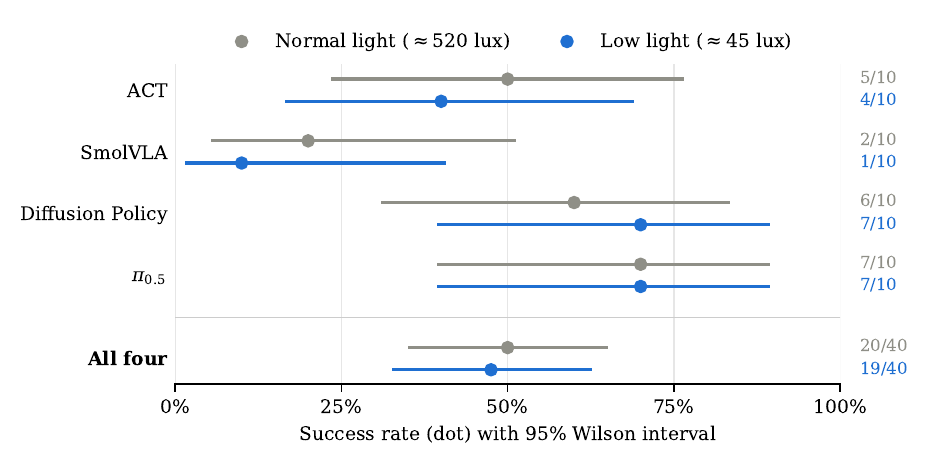}
\caption{Success by policy and light level; 80 rollouts, 10 per cell, placements N00--N09.}
\label{fig:success}
\end{figure}

Every one of the 80 rollouts started. No policy changed by more than one success in ten (Figure~\ref{fig:success}, Table~\ref{tab:effects}).

\begin{table}[h]
\centering\small
\caption{Change in success, low minus normal. Intervals are Newcombe intervals; P(low worse) is the Beta(1,1) posterior. The pooled change is the raw difference; the posterior median of the policy average is $-2.1$ points. All Fisher and McNemar $p$-values were 1.00.}
\label{tab:effects}
\begin{tabular}{lrrr}
\toprule
Policy & Change (pp) & 95\% CI & P(low worse) \\
\midrule
ACT & $-10$ & $-45$ to $+29$ & 0.67 \\
SmolVLA & $-10$ & $-42$ to $+24$ & 0.71 \\
Diffusion Policy & $+10$ & $-28$ to $+45$ & 0.33 \\
\pifive & 0 & $-36$ to $+36$ & 0.50 \\
\midrule
All four & $-2.5$ & $-23$ to $+19$ & 0.59 \\
\bottomrule
\end{tabular}
\end{table}

\begin{itemize}
\item \textbf{Pooled odds ratio} for low light: 0.89 (Mantel--Haenszel, 95\% CI 0.34 to 2.33). The GLM with policy effects gave 0.88 (0.33 to 2.34).
\item \textbf{No evidence that policies differed} in sensitivity (interaction $p = 0.86$; Breslow--Day $p = 0.86$), though the per-policy intervals span about $\pm$40 points.
\item \textbf{A large drop is unlikely:} the posterior probability of a policy-averaged drop of 20 points or more is 0.024. A drop of 10 points or more has probability 0.19 and cannot be excluded.
\item \textbf{No evidence of a shift in failure modes} (permutation $p = 0.77$). Grip failures rose from 6 to 10 across all policies (Fisher $p = 0.40$), mostly in SmolVLA (2 $\to$ 5; Fisher $p = 0.35$). Failure labels are themselves unstable: the two DP low-light runs (Section~\ref{sec:repeat}) gave different labels at the same success count.
\item \textbf{Placements carry some structure:} outcomes matched across conditions at 29 of 40 policy \x{} placement pairs, against 23.6 expected under independence. The placement term in the GLM was $p = 0.033$.
\end{itemize}

\subsection{What the cameras recorded}
When read on 6 Nov after the blocks, both cameras were set to automatic exposure and automatic white balance (Table~\ref{tab:settings}, read with \texttt{v4l2-ctl}). Dynamic frame rate is off by default on both drivers, so something changed these settings at some point before the experiment.

\begin{table}[h]
\centering\small
\caption{Camera settings.}
\label{tab:settings}
\begin{tabular}{lll}
\toprule
Setting & Front (Brio 300) & Wrist (USB 2.0 module) \\
\midrule
Exposure & automatic (aperture priority) & automatic (aperture priority) \\
Dynamic frame rate & on (driver default: off) & on (driver default: off) \\
Gain control & 0 & 0 \\
Backlight compensation & on & on \\
White balance & automatic & automatic \\
Mains flicker filter & 60\,Hz (local mains: 50\,Hz) & 50\,Hz \\
Focus & continuous autofocus & fixed \\
\bottomrule
\end{tabular}
\end{table}

The images show the consequence. Lux at the front and wrist camera positions fell 10.7\x{} and 10.2\x, but mean luma fell only 1.03\x{} and 1.24\x{} (Table~\ref{tab:images}). Luma is gamma-encoded. Converting with a gamma of 2.2, the images received about 7\% and 38\% less linear light: roughly 3\% and 20\% of the drop in log terms, with the rest compensated before the image was encoded.

\begin{table}[h]
\centering\small
\caption{Image statistics, means over episodes (training: 20 S050 episodes).}
\label{tab:images}
\begin{tabular}{lrrrrrr}
\toprule
 & \multicolumn{3}{c}{Front camera} & \multicolumn{3}{c}{Wrist camera} \\
\cmidrule(lr){2-4}\cmidrule(lr){5-7}
Measure & train & normal & low & train & normal & low \\
\midrule
Brightness (mean luma, 0--255) & 157.7 & 161.3 & 155.9 & 163.3 & 161.6 & 130.4 \\
Contrast (luma SD) & 24.8 & 23.5 & 24.9 & 40.3 & 34.3 & 31.6 \\
Sharpness (Laplacian variance) & 37 & 41 & 43 & 493 & 348 & 229 \\
Sharpness / luma$^2$ ($\times10^{-4}$) & 14.8 & 15.7 & 17.5 & 185 & 133 & 134 \\
Saturation & 0.110 & 0.162 & 0.103 & 0.113 & 0.130 & 0.125 \\
Red / green & 1.047 & 1.063 & 1.047 & 1.009 & 1.017 & 1.017 \\
Blue / green & 0.967 & 0.933 & 0.967 & 0.991 & 0.981 & 0.994 \\
Repeated frames while moving & 0.145 & 0.104 & 0.128 & 0.077 & 0.075 & 0.094 \\
\bottomrule
\end{tabular}
\end{table}

\begin{figure}[h]
\centering
\includegraphics[width=0.72\linewidth]{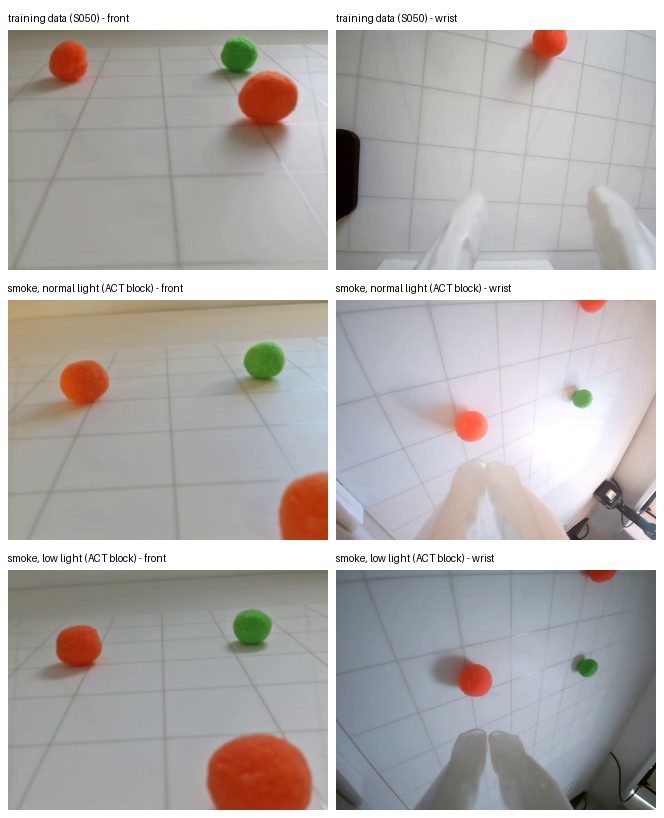}
\caption{First frames of a training episode and of placement N00 in the normal-light and low-light ACT blocks; front camera left, wrist camera right. The low-light frames are barely darker than the others. In normal light the wrist view shows a bright reflection on the floor that does not appear in training or in low light.}
\label{fig:frames}
\end{figure}

\begin{itemize}
\item \textbf{Front camera:} practically unchanged in low light (Figure~\ref{fig:frames}).
\item \textbf{Wrist camera:} 19\% darker and 8\% lower in contrast. Raw sharpness fell 34\%, but that is what dimming alone predicts: normalised for brightness, sharpness did not change (133 vs 134). There is no evidence of added motion blur.
\item \textbf{Frame rate:} repeated frames rose only slightly (front 10\% $\to$ 13\%, wrist 7.5\% $\to$ 9.4\%). A halved frame rate would give about 50\%.
\item \textbf{Colour:} the low-light front images matched the training recordings closely (white balance identical, saturation 0.103 against 0.110). The normal-light images were warmer and 47\% more saturated than training, despite being at the recording lux level. Normal light was itself shifted from training in other ways too: on the wrist camera, contrast (34.3 vs 40.3) and normalised sharpness (133 vs 185) were both below the training recordings.
\item \textbf{Noise:} sensor noise could not be assessed. Video compression removes most grain (the Immerk{\ae}r estimate~\cite{immerkaer} was 0.1--0.5\% of brightness in every condition), whereas the policies saw uncompressed frames.
\item \textbf{Brightness and success:} in low light, episodes with brighter front images succeeded more often ($r = 0.40$ across 40 rollouts, $p = 0.02$). Within policies the association was weaker and not significant (logistic slope 0.28 per grey level, $p = 0.06$). A similar positive slope appears in normal light (0.17, $p = 0.16$), and there is none on the wrist camera ($p = 0.74$). The measure is an episode average, so it also reflects what the episode contained: arm position, ball movement and episode length. It is therefore not a clean exposure variable, and we read the association as a property of episode content rather than evidence of a lighting effect. A cleaner test would use first-frame brightness.
\end{itemize}

\subsection{An unplanned repeat block}\label{sec:repeat}
Re-running the same model in similar light (51 and 59\,lux) reproduced the success count but not which placements succeeded. Through an operator error, the DP low-light command was run a second time, ten minutes after the first DP low-light block. This block is excluded from the comparison above. A planned \pifive{} normal-light repeat, started on 6 Nov at 16:06, was stopped after two complete rollouts and a third of 4.5\,s; it was neither scored nor analysed.

\begin{table}[h]
\centering\small
\caption{The two low-light DP blocks. \checkmark{} success; loc, grip, knock: failure labels.}
\label{tab:repeat}
\setlength{\tabcolsep}{4pt}
\begin{tabular}{lccccccccccc}
\toprule
Placement & N00 & N01 & N02 & N03 & N04 & N05 & N06 & N07 & N08 & N09 & Total \\
\midrule
DP low, run 1 & knock & \checkmark & \checkmark & grip & \checkmark & grip & \checkmark & \checkmark & \checkmark & \checkmark & 7/10 \\
DP low, run 2 & \checkmark & \checkmark & \checkmark & loc & \checkmark & \checkmark & \checkmark & \checkmark & knock & knock & 7/10 \\
\bottomrule
\end{tabular}
\end{table}

The runs agreed at 6 of 10 placements, against 5.8 expected if outcomes were independent ($\varphi = 0.05$). This is consistent with chance, but ten pairs carry little information. Across conditions the 40 policy \x{} placement pairs agreed more often (29/40, $\varphi = 0.45$), and the repeat agreed with DP's normal-light block at 7 of 10. Counting the repeat as extra low-light data leaves the pooled estimate unchanged (odds ratio 0.93, 95\% CI 0.36 to 2.36).

\section{Discussion}
The policies were not shown dark images, so the result is mainly about the camera pipeline: an 11.5\x{} drop in room light reached the front camera's output as a 3\% change in mean luma.

\paragraph{What is and is not shown.} On this rig, as configured, we found no detectable change in success over a range of room light wider than a real bench is likely to drift. Drops of 20 points or more are unlikely; a 10-point drop cannot be excluded. The comparison does not show that ACT, DP, SmolVLA or \pifive{} are robust to darker images: none was tested on them. Equally, the data say nothing about whether vision-language-pretrained encoders (SmolVLA, \pifive) are more robust than an ImageNet-initialised (ACT) or from-scratch (DP) ResNet-18. The image changes that did occur were modest: a 19\% darker wrist view in low light, and front images in normal light that were warmer and more saturated than the training recordings. No policy responded measurably to either.

\paragraph{Relation to earlier work.} Factor decompositions of imitation-learning generalisation found lighting among the easier factors on a real robot~\cite{xie}. Simulated benchmarks rank it among the most damaging perturbations~\cite{colosseum}, while a VLA robustness suite finds it much less damaging than camera-viewpoint changes, and finds that wrist-camera views improve robustness to illumination~\cite{liberoplus}. The difference may lie partly in the camera: simulated perturbations change the rendered image directly, while a real camera's exposure and white-balance control filter a room's light before the policy sees it. Photometric augmentation and domain randomisation~\cite{tobin} make policies robust in training; automatic exposure does part of that job at the sensor.

\paragraph{Noise of a single block.} The accidental repeat shows that one 10-rollout block is a noisy measurement. The same model in similar light, ten minutes apart, gave the same count but matched itself at only 6 of 10 placements: consistent with chance, though ten pairs say little. Placement difficulty adds some structure (29 of 40 matches across conditions), but much of the variation is rollout-to-rollout. Any single-block comparison of one setting against another has to clear that floor.

\paragraph{Implications for real-robot benchmarks.}
\begin{enumerate}
\item \textbf{Report camera settings, not just lux.} Exposure mode, gain, white balance, backlight compensation, autofocus, dynamic frame rate and flicker filter all shape what the policy sees. They are easy to read (\texttt{v4l2-ctl --list-ctrls}) and can silently reset when a camera is reconnected.
\item \textbf{Check images against the training data.} A per-block comparison of brightness, saturation and white balance against the training recordings catches drift that lux misses. Here, normal-light images at the recording lux were further from training in colour than the low-light ones.
\item \textbf{Do not change camera settings between recording and evaluation}, even to fix an apparent mistake such as a 60\,Hz flicker filter on 50\,Hz mains. Doing so is itself a distribution shift.
\item \textbf{To test a policy's own lighting robustness, lock exposure} at the training setting, and say so.
\end{enumerate}

\section{Limitations}
\begin{itemize}
\item \textbf{Small blocks.} Ten rollouts per cell detect only very large per-policy changes; a 70\% $\to$ 40\% drop is detected with probability of about 0.12 (Table~\ref{tab:power}). The pooled estimate is the informative one.
\item \textbf{Condition confounded with day and order.} Normal and low light ran on consecutive days in the same policy order. Light also drifted within each session, and \pifive{} ran about an hour after DP on both days.
\item \textbf{Unblinded, live scoring} by a single operator, with hand placement. Failure labels are subjective and unstable between repeats.
\item \textbf{Floor effect.} SmolVLA succeeded 2/10 in normal light, so it could show at most a 20-point drop.
\item \textbf{No locked-exposure control.} The role of automatic exposure is inferred from the images and the settings, not tested directly.
\item \textbf{Low light changed the light source}, not only its level: artificial light off, daylight screened. Its colour and direction therefore differed as well.
\item \textbf{Camera settings were read after the experiment}, not logged during it. The recorded video is compressed, so sensor noise as seen live is not measured.
\item \textbf{Scope.} One rig, one task, one data level (50 demonstrations), one checkpoint per policy; \pifive{} ran in bf16. Hard shadows, glare, flicker and brighter-than-training light were not tested.
\end{itemize}

\section{Conclusion}
An 11.5\x{} drop in room light did not measurably change the success of ACT, SmolVLA, Diffusion Policy or \pifive{} on an SO-101 pick-and-place task. The pooled change was $-2.5$ points (95\% CI $-23$ to $+19$), with no evidence of a difference between policies. Most of the drop never reached the policies: mean image brightness fell 3\% on the front camera and 19\% on the wrist camera, most likely because both cameras ran automatic exposure. Lux measures the room, not the input. Small real-robot benchmarks should log camera settings and compare image statistics with the training data, and should lock exposure whenever the policy's own robustness is the question.

\section*{Data and code availability}
All ten rollout datasets (the eight comparison blocks, the DP repeat block and the stopped \pifive{} repeat) are public on Hugging Face in LeRobot format, under \nolinkurl{fysical/rollout_s2smoke_*}. The scored rollout log, the analysis scripts (\nolinkurl{analyse_lighting.py} for the statistics, \nolinkurl{image_stats.py} for the image measurements) and their outputs are in the Hugging Face dataset \href{https://huggingface.co/datasets/fysical/so101-lighting-autoexposure}{\nolinkurl{fysical/so101-lighting-autoexposure}}. The model checkpoints belong to a larger study in progress and will be released with it.

\section*{Funding and competing interests}
This work received no external funding. The author is the founder of fysical.ai and declares no competing interests.

\section*{Acknowledgements}
The author designed and ran the experiments, verified all results and takes full responsibility for the content. Claude (Anthropic) was used to assist with code and text drafting.

\appendix
\section{Supplementary tables}

\begin{table}[h]
\centering\small
\caption{Every rollout. \checkmark{} success; loc = location failure; grip = grip failure; knock = ball knocked out of reach; wrong = wrong object. The DP repeat block is in Table~\ref{tab:repeat}.}
\label{tab:all}
\setlength{\tabcolsep}{4pt}
\begin{tabular}{lcccccccc}
\toprule
 & \multicolumn{2}{c}{ACT} & \multicolumn{2}{c}{SmolVLA} & \multicolumn{2}{c}{DP} & \multicolumn{2}{c}{\pifive} \\
\cmidrule(lr){2-3}\cmidrule(lr){4-5}\cmidrule(lr){6-7}\cmidrule(lr){8-9}
Placement & normal & low & normal & low & normal & low & normal & low \\
\midrule
N00 & \checkmark & knock & loc & knock & knock & knock & knock & knock \\
N01 & loc & loc & loc & grip & \checkmark & \checkmark & \checkmark & \checkmark \\
N02 & \checkmark & grip & grip & grip & \checkmark & \checkmark & \checkmark & \checkmark \\
N03 & grip & \checkmark & knock & grip & \checkmark & grip & \checkmark & \checkmark \\
N04 & knock & grip & \checkmark & knock & \checkmark & \checkmark & \checkmark & \checkmark \\
N05 & loc & loc & loc & loc & knock & grip & knock & \checkmark \\
N06 & \checkmark & \checkmark & knock & grip & \checkmark & \checkmark & \checkmark & \checkmark \\
N07 & knock & \checkmark & \checkmark & \checkmark & \checkmark & \checkmark & grip & grip \\
N08 & \checkmark & \checkmark & grip & grip & grip & \checkmark & \checkmark & \checkmark \\
N09 & \checkmark & loc & wrong & wrong & grip & \checkmark & \checkmark & knock \\
\bottomrule
\end{tabular}
\end{table}

\begin{table}[h]
\centering\small
\caption{Illuminance by block (lux), normal / low.}
\label{tab:lux}
\begin{tabular}{lrrrrrr}
\toprule
Block & Workspace & Front camera & Wrist camera & Left & Right & Back \\
\midrule
ACT & 477 / 41 & 505 / 45 & 279 / 25 & 238 / 20 & 423 / 31 & 174 / 19 \\
SmolVLA & 538 / 45 & 513 / 51 & 319 / 38 & 223 / 23 & 333 / 35 & 201 / 24 \\
DP & 506 / 51 & 532 / 55 & 348 / 31 & 251 / 25 & 343 / 27 & 198 / 24 \\
\pifive & 556 / 43 & 583 / 49 & 386 / 36 & 268 / 35 & 333 / 30 & 226 / 27 \\
DP repeat (low only) & 59 & 61 & 35 & 31 & 29 & 30 \\
\bottomrule
\end{tabular}
\end{table}

\begin{table}[h]
\centering\small
\caption{Power of a comparison: probability of detecting a drop (Fisher exact test, two-sided $\alpha = 0.05$; 20{,}000 simulated experiments per row at 10 per condition, 4{,}000 at 40 per condition).}
\label{tab:power}
\begin{tabular}{lrr}
\toprule
True success, normal $\to$ low & 10 rollouts per condition & 40 rollouts per condition \\
\midrule
50\% $\to$ 30\% & 0.05 & 0.36 \\
70\% $\to$ 40\% & 0.12 & 0.72 \\
50\% $\to$ 20\% & 0.13 & 0.75 \\
70\% $\to$ 20\% & 0.43 & 1.00 \\
50\% $\to$ 0\% & 0.62 & 1.00 \\
\bottomrule
\end{tabular}
\end{table}


\begin{thebibliography}{99}\small
\bibitem{act} T.~Z. Zhao, V.~Kumar, S.~Levine, C.~Finn. Learning fine-grained bimanual manipulation with low-cost hardware. RSS 2023; \href{https://arxiv.org/abs/2304.13705}{arXiv:2304.13705}.
\bibitem{dp} C.~Chi, Z.~Xu, S.~Feng, E.~Cousineau, Y.~Du, B.~Burchfiel, R.~Tedrake, S.~Song. Diffusion Policy: visuomotor policy learning via action diffusion. RSS 2023; extended version in IJRR; \href{https://arxiv.org/abs/2303.04137}{arXiv:2303.04137}.
\bibitem{smolvla} M.~Shukor, D.~Aubakirova, F.~Capuano, et al. SmolVLA: a vision-language-action model for affordable and efficient robotics. \href{https://arxiv.org/abs/2506.01844}{arXiv:2506.01844}, 2025.
\bibitem{pi05} Physical Intelligence, K.~Black, N.~Brown, et al. $\pi_{0.5}$: a vision-language-action model with open-world generalization. CoRL 2025, PMLR 305; \href{https://arxiv.org/abs/2504.16054}{arXiv:2504.16054}.
\bibitem{lerobot} R.~Cadene, S.~Alibert, F.~Capuano, M.~Aractingi, et al. LeRobot: an open-source library for end-to-end robot learning. ICLR 2026; \href{https://arxiv.org/abs/2602.22818}{arXiv:2602.22818}.
\bibitem{so101} TheRobotStudio and Hugging Face. SO-ARM100 / SO-101: Standard Open Arm. \url{https://github.com/TheRobotStudio/SO-ARM100}.
\bibitem{xie} A.~Xie, L.~Lee, T.~Xiao, C.~Finn. Decomposing the generalization gap in imitation learning for visual robotic manipulation. ICRA 2024; \href{https://arxiv.org/abs/2307.03659}{arXiv:2307.03659}.
\bibitem{colosseum} W.~Pumacay, I.~Singh, J.~Duan, R.~Krishna, J.~Thomason, D.~Fox. THE COLOSSEUM: a benchmark for evaluating generalization for robotic manipulation. RSS 2024; \href{https://arxiv.org/abs/2402.08191}{arXiv:2402.08191}.
\bibitem{liberoplus} S.~Fei, S.~Wang, J.~Shi, et al. LIBERO-Plus: in-depth robustness analysis of vision-language-action models. \href{https://arxiv.org/abs/2510.13626}{arXiv:2510.13626}, 2025.
\bibitem{tobin} J.~Tobin, R.~Fong, A.~Ray, J.~Schneider, W.~Zaremba, P.~Abbeel. Domain randomization for transferring deep neural networks from simulation to the real world. IROS 2017; \href{https://arxiv.org/abs/1703.06907}{arXiv:1703.06907}.
\bibitem{wilson} E.~B. Wilson. Probable inference, the law of succession, and statistical inference. \emph{J. Am. Stat. Assoc.} 22(158):209--212, 1927.
\bibitem{newcombe} R.~G. Newcombe. Interval estimation for the difference between independent proportions: comparison of eleven methods. \emph{Stat. Med.} 17(8):873--890, 1998.
\bibitem{mh} N.~Mantel, W.~Haenszel. Statistical aspects of the analysis of data from retrospective studies of disease. \emph{J. Natl. Cancer Inst.} 22(4):719--748, 1959.
\bibitem{bd} N.~E. Breslow, N.~E. Day. \emph{Statistical Methods in Cancer Research, Vol.~I: The Analysis of Case-Control Studies.} IARC Sci. Publ. 32, 1980.
\bibitem{immerkaer} J.~Immerk{\ae}r. Fast noise variance estimation. \emph{Comput. Vis. Image Underst.} 64(2):300--302, 1996.
\end{thebibliography}
\end{document}